\documentclass[11pt]{article}

\usepackage{amsmath}
\usepackage{amsfonts}
\usepackage{epsfig}
\usepackage{subfigure}
\usepackage{fancybox}
\usepackage{my-apa-uiuc}

\begin{document}
\sloppy
\begin{titlepage}
\hspace{0.08in}
\begin{minipage}{\textwidth}
\vspace*{2.3in}
\begin{center}
{\bf Decomposable Problems, Niching, \\
and Scalability of Multiobjective\\
Estimation of Distribution Algorithms}\\
\addvspace{0.5in}
{\bf Kumara Sastry}\\
{\bf Martin Pelikan}\\
{\bf David E. Goldberg}\\
\addvspace{0.3in}
IlliGAL Report No. 2005004 \\
February, 2005\\
\vspace*{-0.1in}
\vspace*{3in}
Illinois Genetic Algorithms Laboratory \\
University of Illinois at Urbana-Champaign \\
117 Transportation Building \\
104 S. Mathews Avenue
Urbana, IL 61801 \\
Office: (217) 333-2346\\
Fax: (217) 244-5705 \\
\end{center}
\end{minipage}
\end{titlepage}

\title{Decomposable Problems, Niching, and Scalability of
  Multiobjective Estimation of Distribution Algorithms}
\author{Kumara Sastry$^1$, Martin Pelikan$^2$, David E. Goldberg$^1$\\
~\\
$^1$Illinois Genetic Algorithms Laboratory,\\
    Department of General Engineering,\\ 
University of Illinois at Urbana-Champaign\\
~\\
$^2$Department of Mathematics and computer Science,\\
    University of Missouri at St. Louis\\
~\\
  {\tt ksastry@uiuc.edu, pelikan@cs.umsl.edu, deg@uiuc.edu}
}
\date{}
\maketitle

\begin{abstract}
The paper analyzes the scalability of multiobjective estimation
of distribution algorithms (MOEDAs) on a class of boundedly-difficult
additively-separable multiobjective optimization problems. The paper
illustrates that even if the linkage is correctly identified, massive
multimodality of the search problems can easily overwhelm the nicher
and lead to exponential scale-up. Facetwise models are subsequently
used to propose a growth rate of the number of differing substructures
between the two objectives to avoid the niching method from being
overwhelmed and lead to polynomial scalability of MOEDAs.
\end{abstract}

\section{Introduction}
One of the challenging areas in genetic and evolutionary computation
that has received increased attention is multiobjective evolutionary
algorithms (MOEAs). Several MOEAs have been proposed and applied with
significant success to real-world problems
\cite{Deb:2001:MOGAbook,Coello:2002:MOGAbook}. However, studies on the
theory and analysis of MOEAs have been limited in part because of the
complexity of both the algorithms and the problems. For example, some
aspects of problem difficulty and algorithm scalability have been
recently studied \cite{Deb:1999:MOGAproblems,Chen:2004:PhdThesis}.\par

Recently, there is a growing interest in extending estimation of
distribution algorithms
\cite{Pelikan:2002:EDAsurvey,Larranaga:2002:EDAbook}---a class of {\em
  competent\/} genetic algorithms \cite{Goldberg:1999:Race} that
replace traditional variation operators of genetic algorithms (GAs)
with probabilistic model building of promising solutions and sampling
the model to generate new offspring---to solve multiobjective search
problems quickly, reliably, and accurately. Such multiobjective EDAs
(MOEDAs)
\cite{Bosman:2002:mIDEA,Khan:2002:mBOA,Ocenasek:2002:PhdThesis,Ahn:2005:PhdThesis}
typically combine the model-building and sampling procedures of EDAs
with the selection procedure of MOEAs such as the non-dominated
sorting GA (NSGA-II) \cite{Deb:2000:NSGAII}, and a niching method such
as sharing or crowding in objective space. MOEDAs have been shown to
significantly outperform traditional MOEAs in efficiently searching
and maintaining Pareto-optimal solutions with high probability on
boundedly-difficult problems.\par

However, the scalability of the population size and the number of
function evaluations required by EDAs as a function of problem size
and the number of Pareto-optimal solutions has been largely ignored.
This is the case even though one of the primary motives for designing
MOEDAs is to carry over the polynomial (oftentimes sub-quadratic)
scalability of EDAs to boundedly-difficulty multiobjective search
problems. Therefore, we investigate the scalability of
EDAs---specifically multiobjective extended compact GA (meCGA) and
multiobjective Bayesian optimization algorithm (mBOA)
\cite{Khan:2002:mBOA}---on a class of boundedly-difficult
additively-separable problems. We demonstrate the even if the
sub-structures (or linkages) are correctly identified, massive
multimodality of search problems can overwhelm the niching capability
and lead to exponential scalability. Using facetwise models we predict
the

The paper is organized as follows. We provide a brief background on
the MOEAs, particularly NSGA-II, and MOEDAs, specifically, meCGA. The details on the test problems and experimental methodologies
are described in section~\ref{sec:experiments}.
Section~\ref{sec:results} presents the scalability results of MOEDAs
on a class of boundedly-difficult additively-decomposable
multiobjective problems. Subsequently, we demonstrate how massive
multimodality of the search problem can overwhelm the niching
mechanism and lead to exponential scale-up in
section~\ref{sec:overwhelm}. In section~\ref{sec:circumvent}, using
facetwise models of population-sizing for EDAs and niching methods, we
propose a method to predict the growth rate of the number of
sub-structures to circumvent the nicher from being overwhelmed and
lead to polynomial scalability. Finally, we present key conclusions of
the study.

\section{Background}
In this section we briefly review work on multiobjective evolutionary
algorithms, specifically NSGA-II. We also outline previous work on
multiobjective estimation of distribution algorithms and describe two
MOEDAs that we studied.

\subsection{Multiobjective Evolutionary Algorithms (MOEAs)}
Unlike traditional search methods genetic and evolutionary algorithms
are naturally suited for multiobjective optimization as they can
process a number of solutions in parallel and find all or majority of
the solutions in the Pareto-optimal front. Based on Goldberg's
\cite{Goldberg:1989:GAtext} suggestion of implementing a selection
procedure that uses the non-domination principle, many MOEAs have been
proposed
\cite{Horn:1994:NPGA,Srinivas:1995:NSGA,Fonseca:1993:MOGA,Deb:2000:NSGAII,Zitzler:2002:SPEA2,Corne:2001:PESA2,Erickson:2001:NPGA2,VanVeldhuizen:2000:MOMGA,Zydallis:2001:MOMGA2}.
A detailed survey of various MOEAs is out of the scope of this study
and the interested users should refer to
\cite{Deb:2001:MOGAbook,Coello:2002:MOGAbook} and the references
therein. Since the selection and niching procedure of NSGA-II are used
in the MOEDAs used in this study, we describe them in the following
paragraphs.\par

The selection procedure of NSGA-II consists of three elements:

\noindent
{\bf Non-Dominated sorting:} Which sorts assigns domination ranks to
individuals in the population based on their multiple objective
values. A candidate solution X dominates Y, if X is no worse than Y in
all objectives and if X is better than Y in at lease one objective. In
non-dominated sorting, we start with the set of solutions that are not
dominated by any solution in the population and assign them rank 1.
Next, solutions that are not dominated by any of the remaining
solutions are assigned rank 2. That is, all solutions with rank 2 are
dominated by at least one solution with rank 1, but are not dominated
by others in the population. Thus the sorting and ranking process
continues by assigning increasing ranks to those solutions that are not
dominated by any of the remaining unranked solutions. After
non-dominated sorting, we are left with subsets of population with
different ranks. Solutions with a given rank are not dominated by
solutions that have the same rank or higher and are dominated by at
least one solution with a lower rank. Therefore, with respect to
Pareto optimality, solutions with lower ranks should be given priority.

\par\noindent
{\bf Crowding distance computation:} Apart from finding solutions in
the Pareto front, it is also essential to achieve good coverage or
spread of solutions in the front. The diversity of solutions in the
objective space is usually maintained with a niching mechanism and
NSGA-II uses crowding for doing so. Each solution in the population is
assigned a crowding distance, which estimates how dense the
non-dominated front is in the neighborhood of the solution. Therefore,
the higher the crowding distance of the solution, the more diverse the
solution is in the non-dominated front. The pseudocode for computing
the crowding distance is outlined below:
{\tt
\begin{tabbing}
cro\=wding\_distance\_computation(P)\\
\>for \= rank r = 1 to R\\
\>\>P$_{\texttt{r}}$ = subset of solutions in P with rank r\\
\>\>n$_{\texttt{r}}$ = size(P$_{\texttt{r}}$)\\
\>\>for \= i = 1 to n$_{\texttt{r}}$ \\
\>\>\>d$_{\texttt{c}}$(P$_{\texttt{r}}$(i)) = 0\\
\>\>for \= j = 1 to M\\
\>\>\>Q$_{\texttt{r}}$ = sort P$_{\texttt{r}}$ using j$^{\texttt{th}}$
objective, f$_{\texttt{j}}$.\\
\>\>\>d$_{\texttt{c}}$(Q$_{\texttt{r}}$(1)) = d$_{\texttt{c}}$(Q$_{\texttt{r}}$(n$_{\texttt{r}}$)) = $\infty$\\
\>\>\>for \= i = 2 to n$_{\texttt{r}}$-1\\
\>\>\>\>dist = Q$_{\texttt{r}}$(i+1).f$_{\texttt{j}}$ - Q$_{\texttt{r}}$(i-1).f$_{\texttt{j}}$\\
\>\>\>\>d$_{\texttt{c}}$(Q$_{\texttt{r}}$(i)) = d$_{\texttt{c}}$(Q$_{\texttt{r}}$(i)) + dist\\
\> return d$_{\texttt{c}}$
\end{tabbing}
} 
where, {\tt P} is the population, {\tt R} is the maximum rank
assigned in the population, {\tt M} is the number of objective, and
{\tt Q$_{\texttt{r}}$(i).f$_{\texttt{j}}$} is the value of
j$^{\mathrm{th}}$ objective of the i$^{\mathrm{th}}$ individual.

\par\noindent
{\bf Individual comparison operator:} NSGA-II uses a custom comparison
operator to compare the quality of two solutions and to select the
better individual. Both the rank and the crowding distance of the two
solutions are used in the comparison operator, a pseudo-code of which
is given below. First, the rank of the two individuals are considered
and the solution with a lower rank is selected. If the two individuals
have the same rank, then the solution with the highest crowding
distance is selected.
{\tt
\begin{tabbing}
com\= pare(X,Y)\\
\> if rank(X) < rank(X) then return X\\
\> if rank(X) > rank(Y) then return Y\\
\> if \=rank(X) = rank(Y)\\ 
\>\> if d$_{\texttt{c}}$(X) > d$_{\texttt{c}}$(Y) then return X\\
\>\> if d$_{\texttt{c}}$(X) < d$_{\texttt{c}}$(Y) then return Y\\
\>\> if \=d$_{\texttt{c}}$(X) = d$_{\texttt{c}}$(Y)\\
\>\>\> then randomly choose either X or Y
\end{tabbing}
}

\subsection{Multiobjective Estimation of Distribution Algorithms (MOEDA)}
Similar to single-objective EDAs
\cite{Pelikan:2002:EDAsurvey,Larranaga:2002:EDAbook}, multiobjective
EDAs replace the variation operators of MOEAs with the probabilistic
model building of promising solutions and sampling the model to
generate new offspring. Recently, several MOEDAs have been proposed
\cite{Bosman:2002:mIDEA,Khan:2002:mBOA,Ocenasek:2002:PhdThesis,Khan:2002:Thesis,Bosman:2002:mIDEA1,Laumanns:2002:mBOA,Ahn:2005:PhdThesis}
which have combined variants of the Bayesian optimization algorithm
(BOA) \cite{Pelikan:2000:BOA} and iterated density estimation
of algorithm (IDEA) \cite{Bosman:1999:IDEA,Bosman:2000:mixedIDEA} with
the selection and replacement procedures of MOEAs.\par

Khan {\em et al\/} \cite{Khan:2002:mBOA,Khan:2002:Thesis} proposed
multiobjective BOA (mBOA) and multiobjective hierarchical BOA (mhBOA)
by combining the model building and model sampling procedures of BOA
and hierarchical BOA (hBOA) \cite{Pelikan:2001:hBOA} with the
selection procedure of NSGA-II. They compared the performance of mBOA
and mhBOA with those of NSGA-II on a class of boundedly-difficult
additively-separable deceptive and hierarchically deceptive
functions. Laumanns and Ocenasek
\cite{Laumanns:2002:mBOA,Ocenasek:2002:PhdThesis} combined mixed BOA
with the replacement procedure of strength Pareto evolutionary
algorithm (SPEA2) \cite{Zitzler:2002:SPEA2} and compared it with
NSGA-II and SPEA2 on knapsack problems. Ahn \cite{Ahn:2005:PhdThesis}
combined real-coded BOA with selection procedure of NSGA-II with a
sharing intensity measure and a modified crowding distance metric.\par

Bosman and Thierens \cite{Bosman:2002:mIDEA,Bosman:2002:mIDEA1}
combined IDEAs and mixed IDEAs with non-dominated tournament selection
and clustering. In contrast to other MOEDAs, they used clustering to
split the population into sub-population and separate models were
built for each sub-population. They used the clustering procedure
primarily to obtain a good model of the selected population, but
recently it has been shown that it is one of the essential components
in obtaining a scalable MOGA in general, and MOEDA in particular
\cite{Pelikan:2005:mhBOA}.\par

In this study, we use mBOA and the multiobjective extended compact GA
and test their scalability on a class of boundedly-difficult problems.
The multiobjective extended compact genetic algorithm (meCGA) is
similar to mBOA \cite{Khan:2002:mBOA}, except that the model building
and sampling procedure of BOA is replaced with those of extended
compact GA (eCGA) \cite{Harik:1999:eCGA}. The meCGA is used in this
study in part because the simplicity of the probabilistic model and
its direct mapping to linkage groups makes it amenable to systematic
analysis.  The typical steps of meCGA can be outlined as follows:
\begin{enumerate}
\item {\em Initialization:\/} The population is usually initialized with
  random individuals. However, other initialization procedures can also
  be used in a straightforward manner.
\item {\em Evaluation:\/} The fitness or the quality-measure of the
  individuals are computed.
\item {\em Selection:\/} As in mBOA, we use the selection procedure of
  NSGA-II. That is, we first perform the non-dominated sorting, and
  compute the crowding distance for all the individuals in the
  population. We then use the individual comparison operator to {\em
    bias\/} the generation of new individuals.
\item {\em Probabilistic model estimation:\/} Unlike traditional
  GAs, however, EDAs assume a particular probabilistic model of the
  data, or a {\em class\/} of allowable models. A {\em class-selection
    metric\/} and a {\em class-search mechanism\/} is used to search
  for an optimum probabilistic model that represents the selected
  individuals.\par
 
  \noindent
  {\bf Model representation:} The probability distribution used
  in eCGA is a class of probability models known as marginal product
  models (MPMs). MPMs partition genes into mutually independent
  groups and specifies marginal probabilities for each linkage group.\par

  \noindent
  {\bf Class-Selection metric:} To distinguish between better model
  instances from worse ones, eCGA uses a minimum description length
  (MDL) metric \cite{Rissanen:1978:MDL}. The key concept behind MDL
  models is that all things being equal, simpler models are better
  than more complex ones. The MDL metric used in eCGA is a sum of two
  components:
  \begin{itemize} 
  \item {\bf Model complexity} which quantifies the model representation
    size in terms of number of bits required to store all the marginal
    probabilities:
    \begin{equation}
      \label{eqn:cm}C_m = \log_2(n)\sum_{i = 1}^{m}\left(2^{k_{i}} - 1\right).
    \end{equation}
    where $n$ is the population size, $m$ is the number of linkage
    groups, $k_i$ is the size of the $i^{\mathrm{th}}$ group.
  \item {\bf Compressed population complexity}, which quantifies the data
    compression in terms of the entropy of the marginal distribution
    over all partitions.
    \begin{equation}
      C_p = n\sum_{i = 1}^{m}\sum_{j = 1}^{2^{k_{i}}} -p_{ij}\log_2\left(p_{ij}\right),
    \end{equation}
    where $p_{ij}$ is the frequency of the $j^{\mathrm{th}}$ gene
    sequence of the genes belonging to the $i^{\mathrm{th}}$
    partition.
  \end{itemize}
  
  \noindent
  {\bf Class-Search method:} In eCGA, both the structure and the
  parameters of the model are searched and optimized to best fit the
  data. While the probabilities are learnt based on the variable
  instantiations in the population of selected individuals, a
  greedy-search heuristic is used to find an optimal or near-optimal
  probabilistic model. The search method starts by treating each
  decision variable as independent. The probabilistic model in this
  case is a vector of probabilities, representing the proportion of
  individuals among the selected individuals having a value '{\tt 1}'
  (or alternatively '{\tt 0}') for each variable. The model-search
  method continues by merging two partitions that yields greatest
  improvement in the model-metric score. The subset merges are
  continued until no more improvement in the metric value is possible.
  
\item {\em Offspring creation:\/} New individuals are created by
  sampling the probabilistic model. The offspring population are
  generated by randomly generating subsets from the current
  individuals according to the probabilities of the subsets as
  calculated in the probabilistic model.
  
\item {\em Replacement:\/} We use two replacement techniques in this
  study: (1) Restricted tournament replacement (RTS) \cite{Harik:1995:RTS} in which
  offspring replaces the closest individual among $w$ individuals
  randomly selected from the parent population, only if the offspring
  is better than the closest parent. (2) Elitist replacement used in
  NSGA-II, in which the parent and offspring population are combined. The
  domination ranks and crowding distances are computed on the combined
  population. Individuals with increasing ranks are gradually added
  starting from those with the lowest rank into the new population
  till its size reaches to $n$. However, if it is not possible to add
  all the solutions belonging to a particular rank without increasing
  the population size to greater than $n$, then individuals with
  greater crowding distances are preferred.
 
\item Repeat steps 2--6 until one or more termination criteria are met.
\end{enumerate}

\section{Experiments}\label{sec:experiments}
As mentioned earlier, one of the purposes of this study is to
investigate the scalability of MOEDAs, particularly meCGA and mBOA, on
a class of boundedly-difficult multiobjective problems. In this
section, we describe the test functions used to test the scalability
and the methodology used in obtaining the empirical results.

\subsection{Test Problems}
Our approach in verifying the performance of sub-structural niching is
to consider bounding {\em adversarial problems\/} that exploit one or
more dimensions of problem difficulty \cite{Goldberg:2002:DOI}.
Particularly, we are interested in problems where building-block
identification is critical for the GA success. Additionally, the
problem solver (eCGA) should not have any knowledge of the
building-block structure of the test problem, but should be known to
researchers for verification purposes.\par

One such class of problems is the m-k deceptive {\em trap\/} problem,
which consists of additively separable {\em deceptive\/} functions
\cite{Ackley:1987:deception,Goldberg:1987:deception,Deb:1992:deception}.
Deceptive functions are designed to thwart the very mechanism of
selectorecombinative search by punishing any localized hillclimbing
and requiring mixing of whole building blocks at or above the order of
deception. Using such {\em adversarially\/} designed functions is a
stiff test---in some sense the stiffest test---of algorithm
performance. The idea is that if an algorithm can beat an
adversarially designed test function, it can solve other problems that
are equally hard or easier than the adversarial function.\par

In this study, we use a class of test problems with two objectives:
(1) m-k deceptive trap, and (2) m-k deceptive inverse trap. String
positions are first divided into disjoint subsets or partitions of $k$
bits each. The $k$-bit trap and inverse trap are defined as follows:
\begin{eqnarray}
trap_k(u) &=& 
\left\{
\begin{array}{ll}
1 & \mbox{~~if $u=k$} \\
(1-d)\left[1-{u\over k-1}\right] & \mbox{~~otherwise}
\end{array}
\right.,\\
invtrap_k(u) &=& 
\left\{
\begin{array}{ll}
1 & \mbox{~~if $u=0$} \\
(1-d)\left[{u-1 \over k-1}\right] & \mbox{~~otherwise}
\end{array}
\right.,
\end{eqnarray}
where $u$ is the number of $1$s in the input string of $k$ bits, and
$d$ is the signal difference. Here, we use $k$ $=$ 3, 4, and 5, and
$d$ $=$ 0.9, 0.75, and 0.8 respectively.\par

The m-k trap and inverse trap functions have conflicting
objectives. Any solution that sets the bits in each partition either
to 0s or 1s is Pareto optimal and thus there are a total of $2^m$
solutions in the Pareto-optimal front with $m+1$ distinct points in
the objective space. We investigate the scalability of MOEAs and
consider the population size and number of function evaluations
required to maintain at least one copy of all the representative
Pareto-optimal solutions.\par

To illustrate, how additively decomposable problems with conflicting
objectives can overwhelm the niching mechanism used in
MOEAs---irrespective of linkage adaptation capabilities of the
evolutionary algorithm---and lead to exponential scalability, we
consider a problem where linkage learning is not required.
Specifically, we consider the OneMax-ZeroMax problem which is similar
to bicriteria OneMax problem used by Chen for developing facetwise
models of population sizing and convergence time
\cite{Chen:2004:PhdThesis}. In OneMax-ZeroMax problem, the task is to
maximize two objectives, one which is the sum of all the bits with
value 1, and the other is the sum of all the bits with value 0:
\begin{eqnarray}
f_{{\mathrm{OneMax}}}(X) &=& \sum_{i = 1}^{\ell} x_i,\\
f_{{\mathrm{ZeroMax}}}(X) &=& \sum_{i = 1}^{\ell} (1-x_i),
\end{eqnarray}
where $\ell$ is the problem size, and $x_i$ is the value of the
$i^{{\mathrm{th}}}$ bit of a candidate solution $X$.

\subsection{Methodology}
We tested the scalability of the following MOEAs: (1) Univariate
marginal distribution algorithm (UMDA) \cite{Muhlenbein:1996:UMDA},
(2) NSGA-II with two-point crossover and bit-flip mutation, (3) meCGA,
and (4) mhBOA. For each recombination operator, both elitist
replacement of NSGA-II and restricted tournament replacement were
used.\par

For all test problems and algorithms, different problem sizes were
examined to study scalability. For each problem type, problem size,
and algorithm, a bisection method was used to determine a minimum
population size to allocate at least one individual to each
representative solution in the Pareto front. As mentioned earlier, for
the test problems we consider in this study, for an $\ell$-bit
problem---where $ell$ = $m\cdot k$---there are $2^m$ Pareto-optimal
solutions with $m+1$ distinct objective value pairs. In this study, we
investigate the population size required to (1) find at least one copy of
all the $2^m$ Pareto-optimal solution, and (2) find at least one copy
of the $m+1$ distinct points in the Pareto-optimal front. That is, we
consider Pareto-optimal solutions with the same values of both
objectives to be equivalent.\par

The probability of maintaining at least one copy of all the
representative Pareto-optimal solutions at a given population size is
computed by averaging 10--30 independent MOEA runs. The minimum
population size required to maintain at least one copy of all the
representative solutions in the Pareto front are averaged over 10-30
independent bisection runs. Therefore, the results for each problem
type, problem size, and algorithm correspond to 100--900 independent
GA runs. The number of generations for UMDA, meCGA, and mhBOA was
bounded by $5\ell$, whereas the runs with NSGA-II were given at most
$10\ell$ or $20\ell$ generations because of their slower convergence.

\section{Results}\label{sec:results}
We described the test problems and the experimental methodology used
in testing the scalability of MOEDAs in the previous section. In this
section we present the scalability results, followed by a
demonstration of how multiobjective problems with conflicting
sub-substructures can overwhelm the niching mechanism and lead to
exponential scale-up. Finally, we use facetwise models of population
sizing as dictated by model building, decision making, and supply
\cite{Sastry:2004:eCGAscalability,Pelikan:2003:BOAscalability}, and
niching \cite{Mahfoud:1994:nichingPopSizing} to estimate the growth
rate of conflicting sub-structures that circumvents the niching method
from being overwhelmed and leads to polynomial scalability.

\subsection{Scalability of MOEDAs}
We measure the algorithm performance in terms of minimum number of
function evaluations required to find and maintain at least one copy
of all the representative Pareto-optimal solutions. Even though we
have tried m-k deceptive trap and inverse trap functions for $k$ = 3,
4, and 5, for brevity, we only show results for $k = 3$ in this
paper. However, we note that the results for other values of $k$ are
qualitatively similar and those for $k = 3$ are representative of the
behavior of the MOEAs.\par

\begin{figure}
\center
\subfigure[]{\epsfig{file=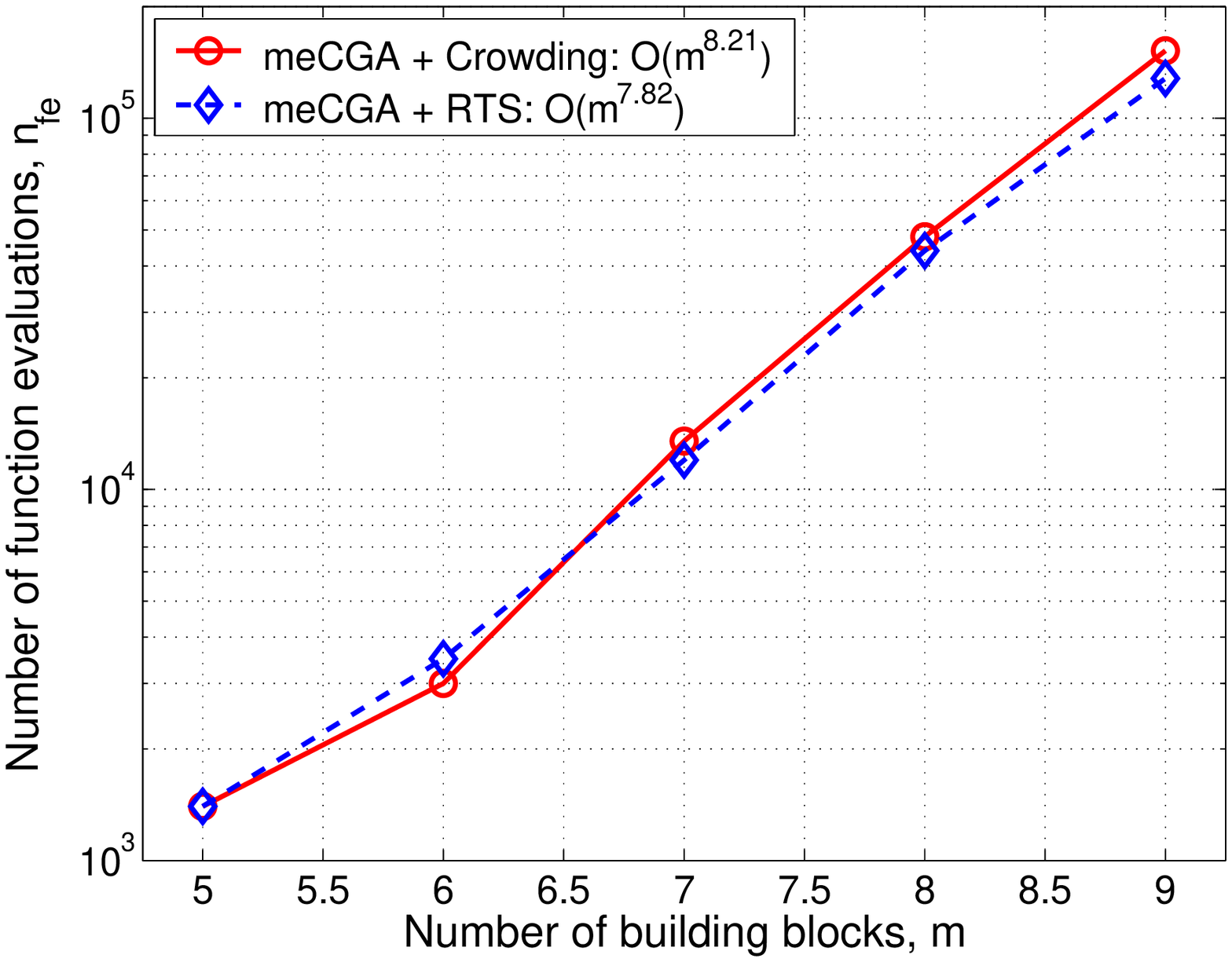,width=3in}}
\subfigure[]{\epsfig{file=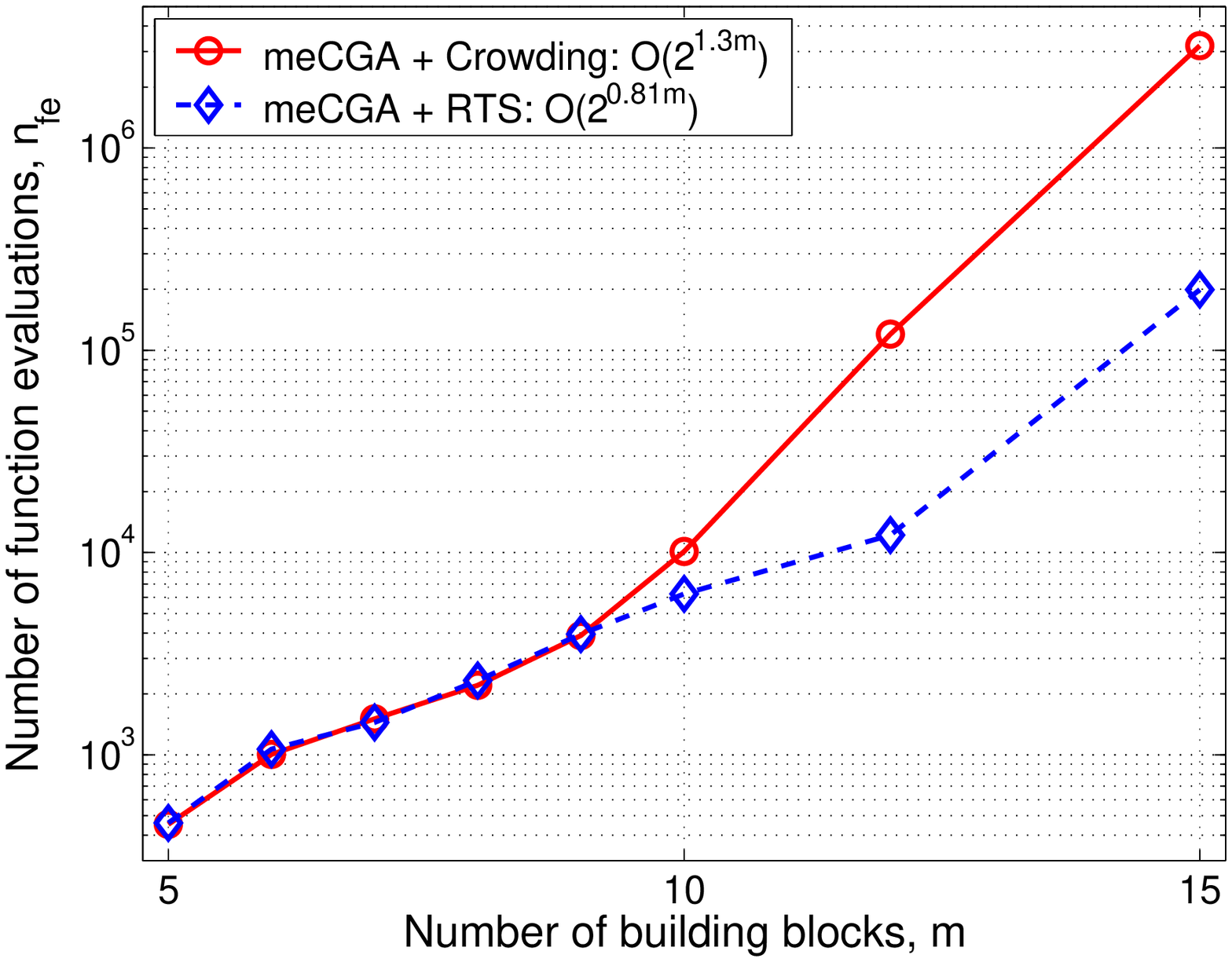,width=3in}}
\caption{Scalability of meCGA with Crowding and with RTS for the m-3 deceptive
  trap and inverse trap with the problem size. Here, we plot the
  minimum number of function evaluations required to search and
  maintain at least one copy of (a) all the $2^m$ solutions in the
  Pareto-optimal front, and (b) only the $m+1$ solutions in the
  Pareto-optimal front with different objective-value pairs. Here, we
  treat the gentotypcially (and phenotypically) different
  Pareto-optimal solution with same values in both objectives to be
  equivalent.}
\label{fig:MOEAscalability}
\end{figure}

Figure~\ref{fig:MOEAscalability}(a), shows the scalability of meCGA
with the problem size for m-3 deceptive trap and inverse trap problem.
We plot the minimum number of function evaluations required to
allocate at least one copy of all the solutions in the Pareto-optimal
front. As shown in the figure, all algorithms scale-up exponentially.
The scale-up does not improve even if we restricted the requirement to
finding only those $m+1$ Pareto-optimal solutions with different
objective-value pairs as shown in Figure~\ref{fig:MOEAscalability}(b).
That is, even if we consider genotypically (and phenotypically)
distinct solutions that have the same value in both objectives to be
equivalent, all the algorithms scale exponentially. This is despite
the linkage information being identified correctly by meCGA and mhBOA
and tight linkage assumption for NSGA-II. Additionally, the
scalability does not improve if the niching or speciation is performed
in the objective space (as in NSGA-II) or in the variable space (as in
restricted tournament selection).\par

Therefore, the exponential scale-up is not due to incorrect linkage
identification and mixing
\cite{Goldberg:1993:mixing,Thierens:1993:mixing,Thierens:1999:mixing},
but because the niching mechanism gets quickly overwhelmed due to the
exponential growth in the number of Pareto-optimal solutions.
Furthermore, the distribution of the $2^m$ solutions in the
Pareto-optimal front is not uniform. There are exponentially as many
solutions in the middle of the front than at the edges (see
table~\ref{tab:nicheDistribution}). That is, there is only one
solution---a binary string with all 0s and all 1s---at each extreme of
the Pareto-optimal front. In contrast, there are
$\left(\begin{array}{c}m\\m/2\end{array}\right) \approx
{\mathcal{O}}\left(e^m\right)$ genotypically different solutions in
the middle of the Pareto-optimal front with same values in both
objectives.

\begin{table}[h]
\center
\begin{tabular}{||l|c|c|c|c|c|c||}\hline\hline
$n_{1,BBs}$  & $0$   & $1$   & $\cdots$ & $i$   & $\cdots$ & $m$\\\hline
$n_{0,BBs}$  & $m$   & $m-1$ & $\cdots$ & $m-i$ & $\cdots$ & $0$\\\hline
\# solutions & 1     & $m$   & $\cdots$ & $\left(\begin{array}{c}m\\i\end{array}\right)$ & $\cdots$ & 1\\\hline\hline 
\end{tabular}
\caption{Distribution of genotypically and phenotypically different
  solutions in the Pareto-optimal front with same values in both
  objectives. $n_{1,BBs}$ refers to the number of $k$-bit partitions
  (substructures) with 1s and $n_{0,BBs}$ is the number of $k$-bit
  partitions with 0s.}
\label{tab:nicheDistribution}
\end{table}

\begin{figure}
\center
\epsfig{file=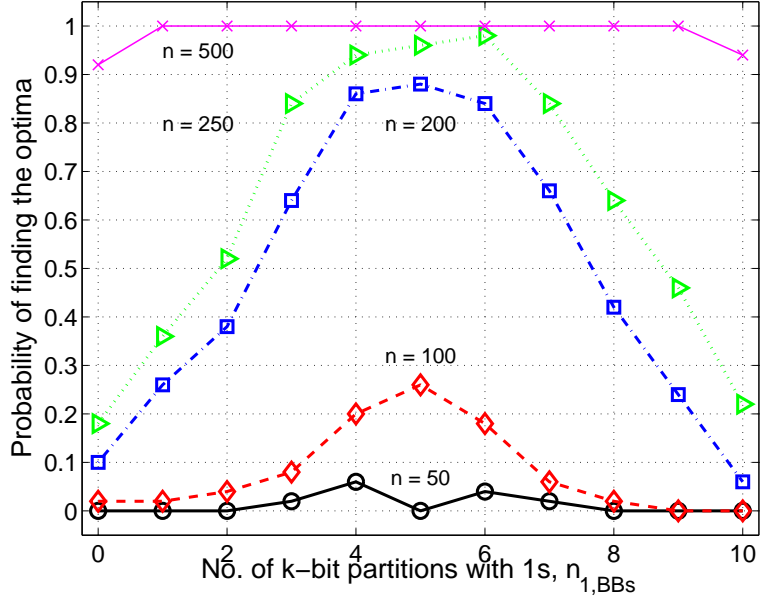,width=4in}
\caption{Probability of finding and maintaining different solutions on
  the Pareto-optimal for the 10-3 deceptive trap and inverse trap
  problem as a function of population size.}
\label{fig:probNiche}
\end{figure}
This highly non-linear distribution of solutions in the Pareto-front
has two effects on the niching mechanisms used in MOEAs in general,
and MOEDAs in particular:
\begin{itemize}
\item Since the extremes of the Pareto-optimal front (maximizing most
  partitions or sub-structures with respect to one particular
  objective) has exponentially smaller representatives than in the
  middle, it takes exponentially longer time, or exponentially larger
  population size \cite{Goldberg:2002:DOI,Thierens:1999:mixing} to
  search and maintain the solutions at the extremes of the
  Pareto-optimal front. When the population size is fixed, the
  probability of maintaining a solution in the middle of the
  Pareto-optimal front is higher than doing so in extremes of the
  front, as shown in figure~\ref{fig:probNiche}. 
\item Since there are multiple points that are genotypically and phenotypically
  different, but lie on the same point on the Pareto-optimal front
  (the solutions have same values in both objectives), some of them
  vanish over time due to drift. The drift affects both the solutions
  in the middle and the near extremes of Pareto-optimal front.
\end{itemize}

\subsection{Overwhelming the Niching Method}\label{sec:overwhelm}
\begin{figure}
\center
\epsfig{file=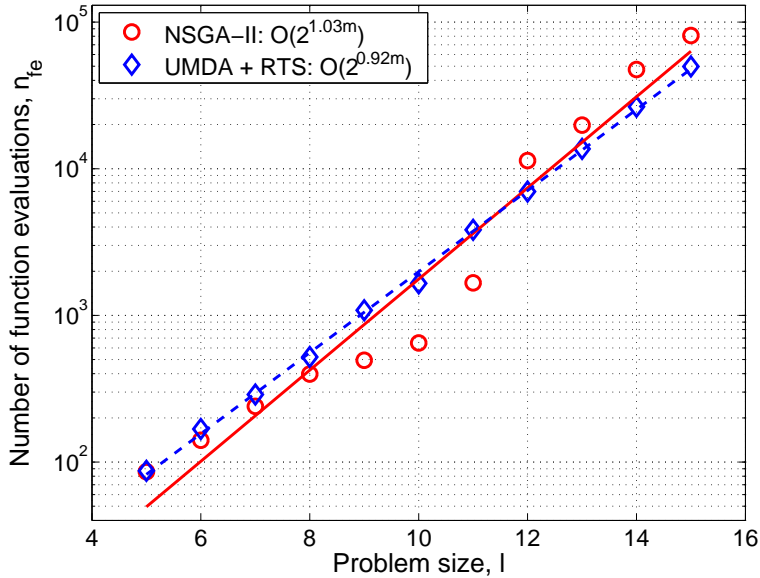, width=4in}
\caption{Scalability of NSGA-II and UMDA on the OneMax-ZeroMax
  problem. Both algorithms with two different niching methods scale-up
  exponentially with the problem size.}\label{fig:nichingOverwhelm}
\end{figure}

To better illustrate how sub-structure competition in all the
partitions of a decomposable problem can lead to nicher overwhelm, we
use the OneMax-ZeroMax problem. We specifically choose the
OneMax-ZeroMax problem to isolate the effect of linkage identification
or lack there of from those of the niching methods on the scalability
of the MOEAs. Unlike the m-k deceptive trap and inverse trap function,
linkage identification is not necessary for the OneMax-ZeroMax
problem. Furthermore, both OneMax and ZeroMax problems are GA-easy
problems which a simple selectorecombinative GA with uniform crossover
and tournament selection can solve in linear time. In contrast, MOEAs
scale-up exponentially in solving OneMax and ZeroMax as shown in
figure~\ref{fig:nichingOverwhelm}.  The results clearly indicate how
the niching methods---both those that work in parameter space (RTS)
and those that work in objective space (Crowding)---get overwhelmed
due to exponentially large number of solutions in the Pareto-optimal
front. Additionally, the results also show that even if the
requirement is relaxed by treating all the different points that lie
on the same point in the Pareto-optimal front to be equivalent, the
scale-up does not improve. Finally, the results suggest that in
decomposable problems, if all or majority of the sub-structures
compete in the two objectives, then the niching method fails to
maintain good coverage, leading to exponential scale-up.

\subsection{Circumventing the Burden on the Niching Method}\label{sec:circumvent}
The results in the previous two sections clearly indicate that MOEDAs
with either RTS or crowding mechanism of NSGA-II scale-up
exponentially with problem size. We also demonstrated that the
exponential scalability is due to the niching method being overwhelmed
because of exponentially large number of solutions in the
Pareto-optimal front. One way to circumvent the niching method from
being overwhelmed is to control the growth rate of the number of
sub-structures that compete in the two objectives, $m_d$. That is, for
a problem with $m$ sub-structures, the two objectives compete in only
$m_d$ sub-structures and share the same $m-m_d$ sub-structures. Since
the total number of Pareto-optimal solutions, $n_{opt} = 2^{m_d}$, by
controlling the number of competing sub-structures, we implicitly
control the total number of Pareto-optimal solutions.\par

The growth-rate of the competing sub-structures should be such that
the effect of model accuracy, decision making, and sub-structure
supply on the population sizing is dominant over the effect of niching
on the population size. The effect of model accuracy, decision making
and sub-structure supply on the population sizing of EDAs is given by
\cite{Pelikan:2003:BOAscalability,Sastry:2004:eCGAscalability}:
\begin{equation}
\label{eqn:eCGApopSizing} n_{{\mathrm{eda}}} \propto c_{1} \cdot 2^{k} \cdot m\log m.
\end{equation}
The effect of niching method on the population-sizing of GAs was
modeled by Mahfoud \cite{Mahfoud:1994:nichingPopSizing} and is
reproduced below:
\begin{equation}
\label{eqn:popSizing} n_{{\mathrm{niching}}} \propto {\log\left[\left(1 -
    \gamma^{1/t}\right)/n_{opt}\right] \over \log\left[\left(n_{opt} -
    1\right)/n_{opt}\right]} \approx c_2 \cdot 2^{m_d},
\end{equation}
where $t$ is the number of generations we need to maintain all the
niches. While Mahfoud derived the population-sizing estimate for
fitness-sharing method, it is generally applicable to other niching
methods and MOEAs as well \cite{Reed:2002:PhdThesis,Khan:2002:Thesis}.

To circumvent the niching method from being overwhelmed we require
$n_{{\mathrm{eda}}} \ge n_{{\mathrm{niching}}}$. That is,
\begin{equation}
c_2 \cdot 2^{m_d} \ge c_{1} \cdot 2^{k} \cdot m\log m.
\end{equation}
The above equation can be approximated \footnote{we neglect the
  $\log_2\left({c_1\log m \over c_2}\right)$ term} to obtain a
conservative estimate of the maximum number of competing
sub-structures that circumvent the niching mechanism from being
  overwhelmed is given by:
\begin{equation}
\label{eqn:growthRate}m_d \approx k + \log_2(m) 
\end{equation}

\begin{figure}
\center
\epsfig{file=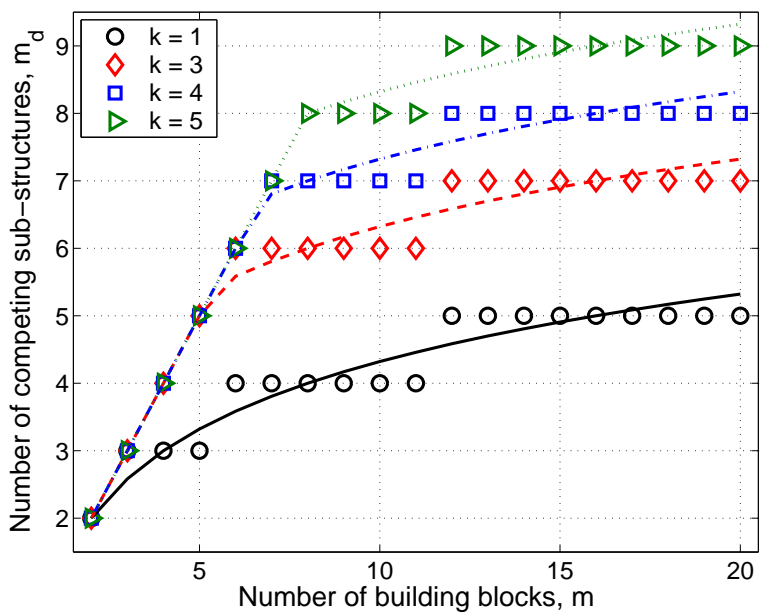, width=4in}
\caption{The growth rate of number of sub-structures that compete in
 the two objectives for different values of $k$ as a function of
 total number of sub-structures.}\label{fig:growthRate}
\end{figure}

The above growth rate is compared to empirical results for different
values of $k$ as a function of total number of sub-structures in the
problem and the results are shown in figure~\ref{fig:growthRate}. As
shown in figure~\ref{fig:growthRate}, the results indicate that once
the growth-rate of competing sub-structures are controlled, the MOEDAs
scale-up polynomially with the problem size, even on the
OneMax-ZeroMax problem. 
\begin{figure}
\center
\epsfig{file=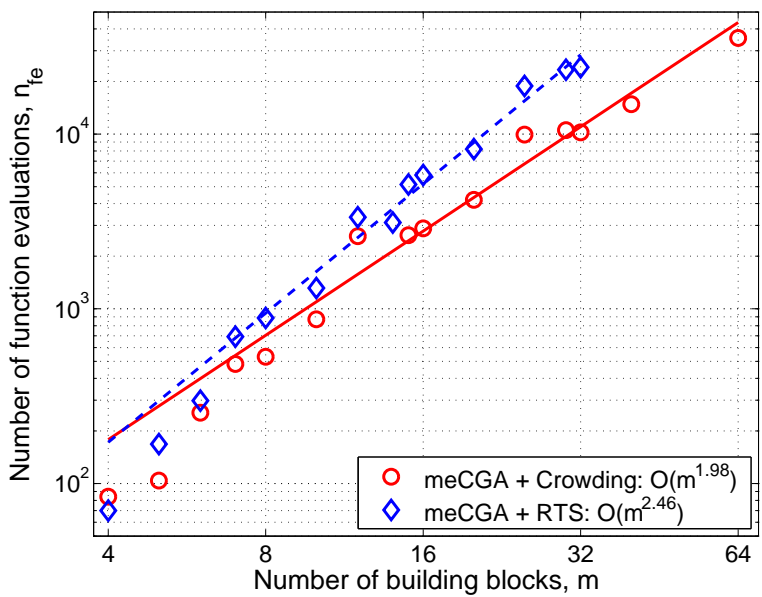, width=4in}
\caption{The scalability of meCGA with the crowding mechanism of
  NSGA-II and RTS niching for both OneMax-ZeroMax and m-3 deceptive
  trap and inverse trap problems. The growth rate of number of
  sub-structures that compete in the two objectives for a given
  problem size is controlled as given by
  equation~\ref{eqn:growthRate}.}\label{fig:growthRate}
\end{figure}

\section{Summary and Conclusions}
In this paper, we studied the scalability of multiobjective estimation
of distribution algorithms (MOEDAs), specifically multiobjective extended
compact genetic algorithm (meCGA) and multiobjective hierarchical
Bayesian optimization algorithm (mhBOA), on a class of
boundedly-difficult additively separable problems. We observed that
even when the linkages were correctly identified, the MOEDAs scaled-up
exponentially with problem size due to failure in the niching
mechanisms. We demonstrated that even if the linkage is correctly
identified, massive multimodality of the search problems can easily
overwhelm the nicher and lead to exponential scale-up. That is, in
decomposable problems, if majority or all the sub-structures compete
in different objectives, then the number of Pareto-optimal solutions
increase exponentially. This exponential increase overwhelms the
nicher and causes significant problems in maintaining a good coverage
of the Pareto-optimal front. Finally, using facetwise models that
incorporate the combined effects of model accuracy, decision making,
and sub-structure supply, and the effect of niching on the population
sizing, we proposed a growth rate of maximum number of sub-structures
that can compete in the two objectives to circumvent the failure of
the niching method. Once the growth-rate of the number of
Pareto-optimal solutions are controlled, the MOEDAs scale-up
polynomially with the problem size.

\section*{Acknowledgments}
This work was sponsored by the Air Force Office of Scientific
Research, Air Force Materiel Command, USAF, under grant
F49620-03-1-0129, the National Science Foundation under ITR grant
DMR-03-25939 at Materials Computation Center, and ITR grant
DMR-01-21695 at CPSD, and the Dept. of Energy under grant
DEFG02-91ER45439 at Fredrick Seitz Materials Research Lab. The U.S.
Government is authorized to reproduce and distribute reprints for
government purposes notwithstanding any copyright notation thereon.

\bibliographystyle{my-apa-uiuc} 
\bibliography{kumaraBibliography}
\end{document}